%% file: main.tex
\documentclass[sigconf, authorversion]{acmart}

\AtBeginDocument{%
  \providecommand\BibTeX{{%
    \normalfont B\kern-0.5em{\scshape i\kern-0.25em b}\kern-0.8em\TeX}}}

\setcopyright{acmcopyright}
\copyrightyear{2018}
\acmYear{2018}
\acmDOI{XXXXXXX.XXXXXXX}

\acmConference[Preprint]{Make sure to enter the correct
  conference title from your rights confirmation emai}{Arxiv}{2022}
\input{setup.tex}

\begin{document}

\title{Progressively-connected Light Field Network\\ for Efficient View Synthesis}



\author{Peng Wang}
\affiliation{
  \institution{The University of Hong Kong}
  \country{Hong Kong, China}
}
\email{pwang3@cs.hku.hk}

\author{Yuan Liu}
\affiliation{
  \institution{The University of Hong Kong}
  \country{Hong Kong, China}
}
\email{yliu@cs.hku.hk}

\author{Guying Lin}
\affiliation{%
  \institution{Zhejiang University}
  \country{China}
}
\email{3180103404@zju.edu.cn}

\author{Jiatao Gu}
\affiliation{%
 \institution{Apple*\thanks{*Work done at Meta AI}}
 \country{United States}
}
\email{thomagram@gmail.com}

\author{Lingjie Liu}
\affiliation{
  \institution{Max Planck Institute for Informatics}
  \country{Germany}    
}
\email{lliu@mpi-inf.mpg.de}

\author{Taku Komura}
\affiliation{%
  \institution{The University of Hong Kong}
  \country{Hong Kong, China}
}
\email{taku@cs.hku.hk}

\author{Wenping Wang}
\affiliation{%
  \institution{Texas A\&M University}
  \country{United States}
}
\email{wenping@tamu.edu}

\renewcommand{\shortauthors}{}

\input{0_abstract}


\input{figures/fig_teaser}

\maketitle
\input{1_intro}

\input{2_related_works}
\input{3_method}
\input{4_experiments}
\input{5_conclusion}
\input{6_ack}


\bibliographystyle{ACM-Reference-Format}
\bibliography{reference}

\clearpage
\setcounter{section}{0}
\renewcommand{\thesection}{\Alph{section}} 
\newpage

\begin{center}
{ \bf\LARGE - ProLiF: Supplementary -}
\end{center}

\input{supp_1_addition_method_descriptions}

\input{supp_2_additional_results}


\end{document}

%% file: setup.tex
\usepackage{multirow}
\usepackage{multicol}
\usepackage{listings}
\usepackage{xcolor}

\definecolor{codegreen}{rgb}{0,0.6,0}
\definecolor{codegray}{rgb}{0.5,0.5,0.5}
\definecolor{codepurple}{rgb}{0.58,0,0.82}
\definecolor{backcolour}{rgb}{0.95,0.95,0.92}

\lstdefinestyle{mystyle}{
    backgroundcolor=\color{backcolour},   
    commentstyle=\color{codegreen},
    keywordstyle=\color{magenta},
    numberstyle=\tiny\color{codegray},
    stringstyle=\color{codepurple},
    basicstyle=\ttfamily\footnotesize,
    breakatwhitespace=false,         
    breaklines=true,                 
    captionpos=b,                    
    keepspaces=true,                 
    numbers=left,                    
    numbersep=5pt,                  
    showspaces=false,                
    showstringspaces=false,
    showtabs=false,                  
    tabsize=2
}

%% file: 0_abstract.tex
\begin{abstract}

This paper presents a \textbf{Pro}gressively-connected \textbf{Li}ght \textbf{F}ield network (\textbf{ProLiF}), for the novel view synthesis of complex forward-facing scenes. ProLiF encodes a 4D light field, which allows rendering a large batch of rays in one training step for image- or patch-level losses. 
Directly learning a neural light field from images has difficulty in rendering multi-view consistent images due to its unawareness of the underlying 3D geometry. 
To address this problem, we propose a progressive training scheme and regularization losses to infer the underlying geometry during training, both of which enforce the multi-view consistency and thus greatly improves the rendering quality. Experiments demonstrate that our method is able to achieve significantly better rendering quality than the vanilla neural light fields and comparable results to NeRF-like rendering methods on the challenging LLFF dataset and Shiny Object dataset. Moreover, we demonstrate better compatibility with LPIPS loss to achieve robustness to varying light conditions and CLIP loss to control the rendering style of the scene. Project page: \href{https://totoro97.github.io/projects/prolif}{https://totoro97.github.io/projects/prolif}.


\end{abstract}

%% file: figures/fig_teaser.tex
\begin{teaserfigure}
  \includegraphics[width=\textwidth]{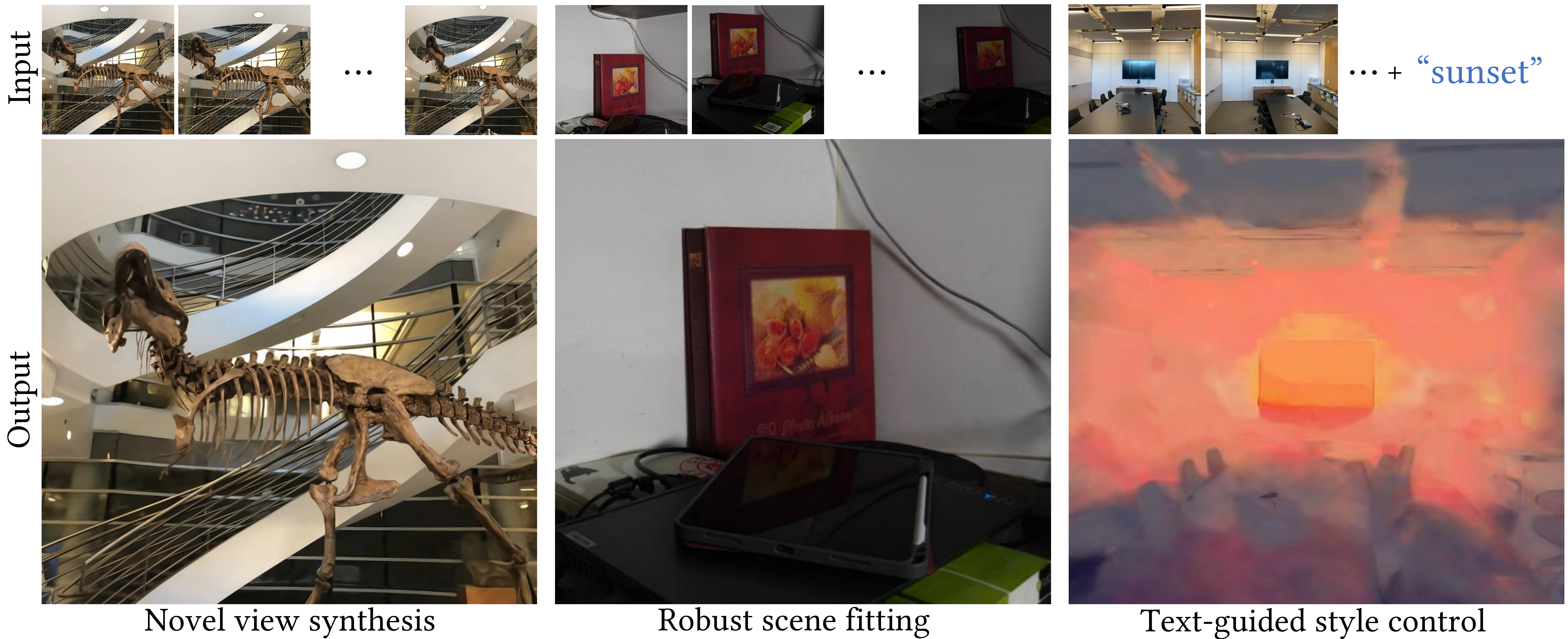}
  \caption{We present ProLiF, a compact neural light field representation for efficient photo-realistic novel view synthesis (left). ProLiF has good compatibility with image-level losses such as LPIPS~\cite{zhang2018perceptual} loss to achieve robustness to varying lighting conditions of input images (middle), and CLIP loss~\cite{radford2021learning} for multi-view consistent scene style editing (right).}
  \label{fig:teaser}
\end{teaserfigure}

%% file: 1_intro.tex
\section{Introduction}
Rendering free-viewpoint photo-realistic images from camera captures is a long-standing problem in computer vision, multimedia and graphics. This task, which we usually call image based rendering (IBR) or novel view synthesis (NVS), has many practical applications such as AR/VR, human-computer interaction, etc. 

Recently, the rendering quality of IBR has been improved drastically with the developments of Neural Radiance Fields (NeRF)~\cite{mildenhall2020nerf}, which represents a scene as a radiance field encoded by the Multi-layer Perceptrons (MLPs) and renders images of the scene by the differentiable volume rendering on the field. Training of NeRF is mainly based on a pixel-level loss that compares the rendered color from the NeRF with the ground-truth color pixel-by-pixel. 
Recent works have shown that training a NeRF model with advanced patch- or image-level losses like GAN loss~\cite{schwarz2020graf,niemeyer2021giraffe,gu2021stylenerf} or CLIP loss~\cite{radford2021learning,jain2021zero,wang2021clip,hong2022avatarclip} would greatly expand the scope of NeRF applications in image generation or image editing.
However, computing patch- or image-level losses requires rendering a large batch of pixels in one training step. 
Since NeRF requires hundreds of network evaluations to synthesize a pixel color, these recent works either only synthesize small patches by NeRF or need multiple high-end GPUs with rather large device memories to compute such high-level losses in training. 
When rendering a large batch, these methods cost excessively-long rendering time and huge GPU memory. 


To improve the efficiency, recent works~\cite{liu2020neural,yu2021plenoctrees,reiser2021kilonerf,hedman2021baking,garbin2021fastnerf,piala2021terminerf,fang2021neusample,neff2021donerf} take advantage of voxels~\cite{liu2020neural,yu2021plenoctrees,reiser2021kilonerf,hedman2021baking,garbin2021fastnerf} or networks~\cite{piala2021terminerf,fang2021neusample,neff2021donerf} to memorize the surface distributions and only sample few points near the surfaces for rendering.
However, these speed-up methods are only applicable in the inference stage or in the late training stage after convergence to valid surfaces, but not at the beginning of the training.
It is because they require a good estimation of the surfaces to guide the sampling of points for rendering acceleration.

\input{figures/fig_overfit}
Instead of learning a radiance field, some recent works manage to learn a neural {\it light field}, that directly maps a ray representation to a color~\cite{sitzmann2021light,liu2021neulf,attal2021learning}.
This direct prediction avoids exhaustive evaluations of the network like NeRF, and only need one forward pass to synthesize a pixel color in both training and inference stages, which greatly improves the rendering efficiency and reduces the memory consumption.
Hence, neural light fields are more suitable to be trained with image- or patch-level supervision.
The main limitation of neural light fields is the lack of multi-view consistency in generating different views of the same scene, because such neural light fields are 3D geometry-agnostic. For example, for two different rays, it is difficult for the light field network to know whether these two rays intersect to the same 3D surface point so that the rendered colors for these two rays should be consistent. In this case, neural light fields easily overfit to input images 
rather than representing the correct scene geometry, leading to severe artifacts in the novel rendered views as shown in Fig.~\ref{fig:overfit}.


To address these problems, we propose a Progressively-connected Light Field network, called ProLiF,
that preserves multi-view consistency in the rendering and enables fast rendering in both training and inference stages without extensive memory consumption.
Instead of directly learning a neural network to map a ray to a color, we design a point-based neural light field. In our method, we evenly sample points on the ray and use the concatenation of all point coordinates as the parameterization of the ray. We then feed these coordinates into a network to predict the density values and the colors of all these sample points in a single forward pass. Finally, we accumulate the colors and density values in a volume rendering fashion to compute the final pixel color.

Simply applying the point-based neural light field still cannot perfectly keep multi-view consistency, i.e., the generalization ability to novel views, because the prediction of the color and the density on a sample point will be entangled with coordinates of other points on the same ray. 
Therefore, we additionally propose a progressive training scheme and multi-view consistency losses, to further improve the generalization ability. In the progressive training scheme, the points along the ray are sampled, and the network with sparse neuron connections processes every point separately at the beginning of the training, which enables a coarse but geometry-consistent initialization of the scene representation. Then, in the subsequent training steps, the neuron connections inside the network are gradually increased to improve the capacity of our neural representation for high-quality rendering.
Meanwhile, during training, the novel multi-view consistency losses are used, which force rays intersecting on a surface point to have consistent colors.


We have conducted experiments on the LLFF dataset~\cite{mildenhall2019llff} and the Shiny Object dataset~\cite{wizadwongsa2021nex} to demonstrate the effectiveness of our method. Results show that the proposed method achieves better quality than the baseline neural light fields and comparable rendering quality to NeRF~\cite{mildenhall2020nerf}.
We also demonstrate that our method is well-compatible with LPIPS loss~\cite{zhang2018perceptual} to achieve robustness to different lighting conditions, and with CLIP loss to manipulate rendering style using texts. Moreover, our neural light field can efficiently render images at the resolution of $400 \times 400$ at 5 frames-per-second (FPS) using a model size of 11MB.

Our contributions can be summarized as:
\begin{itemize}

  \item A novel and compact neural light field representation for fast view synthesis in the both training and inference time without extensive memory consumption for a complex scene. The fast rendering also enables the use of the high-level losses on the full image during training. 
  \item A progressive training scheme that provides good trade-off between the network capacity and multi-view consistency.
  \item Novel regularization losses for preserving multi-view consistency in the rendering of the neural light field.
\end{itemize}

%% file: figures/fig_overfit.tex
\begin{figure}[htb]
  \includegraphics[width=\linewidth]{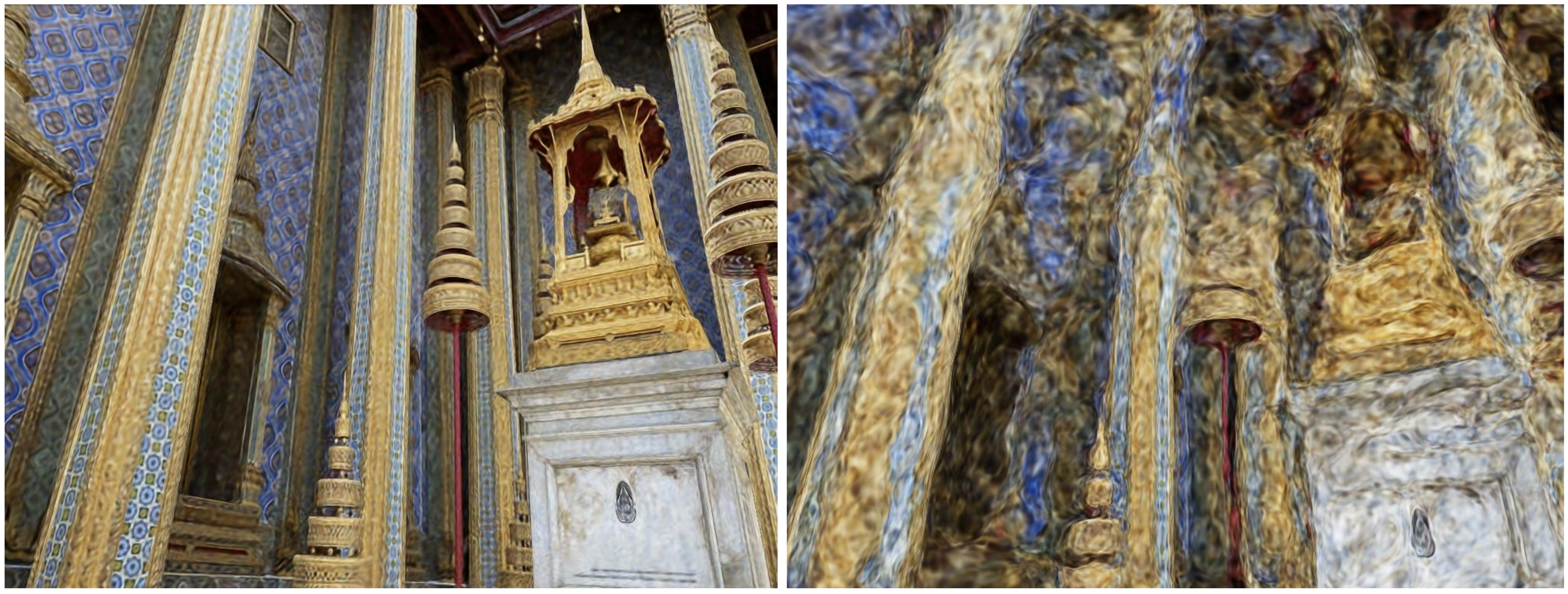}
  \caption{A vanilla neural light field network is able to fit the training images (left) but cannot generalize to novel views (right). Here we use the LFN~\cite{sitzmann2021light} with SIREN activations~\cite{sitzmann2020implicit} to fit the 4D light field. }
  \label{fig:overfit}
\end{figure}

%% file: 2_related_works.tex
\section{Related Works}

\textbf{Traditional IBR}. Novel view synthesis is a long-standing problem. Previously, novel view synthesis is mainly solved by blending input images~\cite{levoy1996light, chaurasia2013depth, kalantari2016learning, hedman2016scalable, kopanas2021point, hedman2018deep, choi2019extreme, thies2020ignor, xu2019deep, Riegler2020FVS, Riegler2021SVS} or fitting a plenoptic function~\cite{gortler1996lumigraph, davis2012unstructured}. The key problem to these methods is to select the correct combination of input colors to blend the image or form the plenoptic color. To address this problem, these methods either require very dense input views~\cite{gortler1996lumigraph, levoy1996light} or rely on an external geometry reconstruction~\cite{chaurasia2013depth, penner2017soft}. Our method can be regarded as a neural plenoptic function which uses a neural network to map input rays to colors by fitting the input images. We do not resort to external geometry reconstruction but apply progressive training scheme and regularization to help the network learn the geometry.

\vspace{2pt}
\noindent{\textbf{Neural Radiance field}}. Recent works~\cite{sitzmann2019deepvoxels, lombardi2019neural,thies2019deferred, liu2019neural, liu2020NeuralHumanRendering,habermann2021,liu2021neuralactor,wu2020multi, aliev2020neural,ruckert2021adop,kopanas2021point,sitzmann2019scene,mildenhall2020nerf,liu2020neural,peng2021neural,liu2020dist,niemeyer2020differentiable,Kellnhofer2021nlr,wang2021neus,liu2021neural} show that we are able to learn neural scene representations for the NVS task. Among these works, NeRF~\cite{mildenhall2020nerf} demonstrates significant improvements on this task by learning a neural radiance field from input images. However, NeRF still suffers from insufficient rendering in both training or inference stages. Many follow-up works proposes voxels~\cite{yu2021plenoctrees,liu2020neural,hedman2021baking} or other network designs~\cite{lindell2021autoint,neff2021donerf} to speed up the rendering in inference stage.
However, these methods do not allow efficiently rendering a large batch of rays in the training stage while our method can do this for better compatibility with image- or patch-level losses like LPIPS or CLIP losses.

\vspace{2pt}
\noindent{\textbf{Neural Light field}}. Light Field Network (LFN)~\cite{sitzmann2021light,smith2022unsupervised} shows the possibility to learn a neural light field to directly map rays to colors for the NVS task. However, training a neural light field easily overfits on input images due to the absence of an explicit 3D geometry. Concurrent works~\cite{suhail2021light, attal2021learning} solves this problem by adding features of sample points along rays~\cite{suhail2021light} or learning subdivided neural light fields~\cite{attal2021learning}, which achieve better rendering quality at a cost of more computation in training and thus does not allow rendering a large batch of rays in one training step. The most similar concurrent work is NeuLF~\cite{liu2021neulf} which applies a depth-based regularization to prevent overfitting. Our method uses a progressive training scheme and different regularization losses, which demonstrates better rendering quality than NeuLF.

\vspace{2pt}
\noindent\textbf{High-level losses for NeRF}. Some recent works managed to apply high-level losses on NeRF, like GAN loss~\cite{schwarz2020graf,niemeyer2021giraffe,chan2021pi,gu2021stylenerf} or CLIP loss~\cite{jain2021putting,wang2021clip,jain2021zero,hong2022avatarclip}. These works use CLIP loss for sparse view reconstruction~\cite{jain2021putting}, generation~\cite{jain2021zero, hong2022avatarclip} and style editing~\cite{wang2021clip}. Among them, CLIP-NeRF~\cite{wang2021clip} also intends for style editing on NeRF but uses an advanced conditional NeRF pre-trained on large training set. In contrast, our method does not require pre-training and enables rendering a large batch rays in on training step.

%% file: 3_method.tex
\input{figures/fig_preliminary}
\input{figures/fig_arch}

\section{Method}
Given a set of input images with known camera poses, our goal is to learn a neural light field from these images. 
Assuming all cameras capturing the input images are forward-facing to the scene, rays emitted from these images can be parameterized by $r=(u, v, s, t)$ as shown in Fig.~\ref{fig:preliminary} (a), where $(u, v)$ is the point coordinate of the near plane and $(s, t)$ is that of the far plane. The targeted neural light field is a function $LF: \mathbb{R}^4\rightarrow \mathbb{R}^3$ that maps the ray $r$ to a color $c\in \mathbb{R}^3$. A straightforward way to model this function is to train an MLP that maps $(u,v,s,t)$ to the color $C$ on input images.
However, such a design is 3D geometry-agnostic, which easily overfits to input images without the ability to generalize to novel views as shown in Fig.~\ref{fig:overfit}. 
To enable the neural light field to learn a multi-view consistent scene representation, we propose a point-based neural light field (Sec.~\ref{sec:point}), a progressive training scheme (Sec.~\ref{sec:progress}), and a multi-view consistency regularization scheme (Sec.~\ref{sec:regularization}) to improve its generalization ability to novel views.

\subsection{Point-based neural light field}
\label{sec:point}

On the given ray, we \textit{evenly} sample points at predefined depth values $\{d_0, d_1, ..., d_{D-1}\}$ between the near and the far plane, as shown in Fig.~\ref{fig:preliminary} (b). Then, the ray is paramterized by the concatenation of all point coordinates $(x_0,y0,x_1,y_1,...,x_{D-1},y_{D-1})$ and is fed into a Multi-layer Perceptron $MLP:\mathbb{R}^{2D} \to \mathbb{R}^{D}\times \mathbb{R}^{3D}$ to predict densities and colors at all these points $\{(\sigma_i,c_i)|i=0,...,D-1\}$ in one forward pass. Finally, similar to NeRF~\cite{mildenhall2020nerf}, the colors and densities at these points are accumulated by a volume rendering scheme to compute the rendered color $C$ for this ray as follows:
\begin{equation}
    C = \sum_{i=0}^{D-1} T_i{\alpha_i} c_i,
\end{equation}
where $\alpha_i$, $T_i$ are the alpha value and the transmittance on the $i$-th point respectively, which are computed by
\begin{equation}
\begin{split}
    \alpha_i &= 1 - \exp(-\delta_i\sigma_i) \\
    T_i &= \prod_{j=0}^{i-1} (1 - \alpha_j).
    \label{eqn:volume_rendering}
\end{split}
\end{equation}
Here $\delta_i$ denotes the distance between $i$-th and $(i+1)$-th point. 

\textbf{Discussion}. In contrast to the straightforward design that directly maps a ray to a color, our neural light field predicts densities and colors at the sampled points and 
the final pixel color is then computed by volume rendering. This scheme provides the network an opportunity to infer the occlusion relationships among points and to learn the underlying scene geometry. Meanwhile, in comparison with NeRF~\cite{mildenhall2020nerf} which needs to forward-pass the network multiple times with all sample points, our network predicts all densities and colors in a single forward-pass of the MLP. Thus, our method is much more efficient and allows rendering a large batch of rays in one training step. 

However, our efficient computation comes at a cost that the color and the density of a 3D point are not only dependent on its own coordinate but also entangled with coordinates of other point samples on the same ray. 
In this case, when two rays intersect on a 3D point, the neural light field possibly assigns totally different colors and densities for this 3D point on two rays, which leads to inconsistency in scene geometry and bad generalization to novel views. To tackle this issue, in the following, we introduce the progressive training scheme and multi-view consistency regularizations in the following.

\subsection{Progressive training scheme}
\label{sec:progress}
Considering that the inconsistency mainly comes from the entanglement between point coordinates on a ray, we propose a progressive training scheme to process every point independently at the beginning of training to learn consistent scene geometry and then gradually improve the capacity of our network for better rendering quality by adding connections in the network.

Our key idea is that at the beginning of training, we discard parts of the neuron connections in our MLP network to ensure that the color or density of a point only depends on the coordinate of this point without being entangled with coordinates of other points. 
In this case, a point will have the same color and density on different rays, which provides a strong inductive bias for the network to learn a consistent scene geometry.
However, to ensure small memory consumption and efficient computation, the MLP network at the beginning is small with sparse connections, whose capacity is not enough to build a complex scene representation for high-quality rendering. Thus, in the subsequent training, we gradually increase the connections in our MLP to improve the capacity.

Specifically, as shown in Fig.~\ref{fig:arch}, we split the training into $k$-stages. At the beginning stage 0, the coordinate of a point at a specific depth only connects to its own feature vector thus it does not affect the features of points at other depth samples. In stage 0, given $D$ depth samples, this architecture can be regarded as $D$ separated small subnetworks which process the points at different depth independently to predict the colors and densities. Note that points at the same depth on different rays share the same small subnetwork. After training on stage 0 for $t$ steps, we switch the training to the next stage. 
During switching, two neighboring subnetworks will be merged into one single fully connected network by concatenating features of the two subnetworks and adding connections between adjacent layers of the merged network. In stage 1, we will have $D/2$ subnetworks, each of which processes coordinates of two neighboring point samples and produces the colors and densities for these two points.
The new connections are constructed by merging weight parameters of the subnetworks and adding zero values to the new weight parameters. 
Note in the merging, the total channel number in the hidden layers of all points stays unchanged.
After switching, we will continue training with the architecture of the stage 1 for $t$ steps. Then, we repeat the merging and training until we get a single full-connected network to process all input point coordinates.
More details can be found in the supplementary materials.

\vspace{2pt}
\noindent{\bf Progressive depth sampling.} Besides increasing network capacity by adding more neuron connections, we also begin with a very sparse depth samples between the near plane and far plane (Fig. \ref{fig:preliminary}), and progressively add more sampling depth values in the training. In order to process points at newly-added depth samples, we interpolate weights in subnetworks of neighboring depth values. For more details, please refer to supplementary materials.

\subsection{Multi-view consistency regularization}
\label{sec:regularization}
In the whole training process, we apply multi-view consistency regularizations to force the network to learn a consistent underlying scene geometry.

\vspace{2pt}
\noindent{\bf Density consistency loss.}
The density consistency loss is based on the fact that the density of a 3D point is a constant value regardless which the input ray the point locates. However, in the later training stages, the density of a point would also be affected by other points on the ray. Thus, the predicted densities of the same point on different rays would be different. This motivates us to design a loss to constrain the predicted densities of the same point on different rays to be consistent. In the following, we derive a loss to implement this density constraint.

To derive the loss, we denote the whole process of computing a density $\sigma_i$ on a point $p_i$ at depth $d_i$ on the ray $(u,v,s,t)$ as $\sigma_i=L_i(u,v,s,t)$. Note $L_i$ consists of (1) sampling points on the ray $(u,v,s,t)$ with predefined depth values and (2) predicting the density value at all the sample points. Meanwhile, $L_i$ is differentiable with $(u, v, s, t)$
 because the both suboperations are differentiable.
The density consistency loss for this point at depth $d_i$ is
\begin{equation}
    \mathcal{L}_{\rm density} = \mathbb{E}_{u,v,s,t,i}\left(\|\nabla L_i(u, v, s, t)\cdot \mathbf{p}\|^2 + \|\nabla L_i(u, v, s, t)\cdot \mathbf{q} \|^2 \right),
\end{equation}
where
\begin{equation}
\begin{split}
    \mathbf{p} &= (d_i, 0, -(1 - d_i), 0)^{\top} \\
    \mathbf{q} &= (0, d_i, 0, -(1 - d_i))^{\top}.
\end{split}
\end{equation}
The intuition behind this loss is that (1) all rays intersecting the point $p_i$ form a 2D manifold in $(u,v,s,t)$; (2) $(\mathbf{p},\mathbf{q})$ is a basis spanning the tangent space of this manifold on this point; (3) we minimize the gradient $\nabla L_i$'s projection in this tangent space.  Ideally, the densities of this point will be the same for all rays intersecting on this point so such a projection will be zero. We provide detailed derivation in the supplementary material.

\input{figures/fig_compare_llff}

\vspace{2pt} 
\noindent{\bf Color consistency loss.} 
Similarly, we expect the rendered colors for rays intersecting the same surface point to have consistent colors. We follow the similar process to represent the whole rendering procedure of our neural radiance field by $C=LF(u,v,s,t)$ and compute a color consistency loss by
\begin{equation}
    \mathcal{L}_{\rm color} = \mathbb{E}_{u,v,s,t}\left(\mathcal{R}(\nabla LF(u, v, s, t)\cdot \mathbf{p}) + \mathcal{R}(\nabla LF(u, v, s, t)\cdot \mathbf{q})\right),
\end{equation}
where
\begin{equation}
\begin{split}
    \mathbf{p} &= (\hat{d}, 0, -(1 - \hat{d}), 0)^{\top} \\
    \mathbf{q} &= (0, \hat{d}, 0, -(1 - \hat{d}))^{\top} \\
    \hat{d} &= \sum_{i} T_i\alpha_i d_i.
\end{split}
\end{equation}
Here $T_i$ and $\alpha_i$ are the transmittance and the alpha values in Eqn.~\ref{eqn:volume_rendering}, $\hat{d}$ is the predicted depth of the visible surface point for this ray, and $\mathcal{R}$ is chosen as smooth L1 loss~\cite{girshick2015fast} to reduce sensitivity to outliers on occlusion boundaries. The intuition behind this color consistency loss is similar to the density consistency loss, which computes the tangent space basis $(\mathbf{p},\mathbf{q})$ at the depth $\hat{d}$ and minimizes the projection of $\nabla LF$'s gradient in this tangent space.


In summary, the total loss to train our neural light field is
\begin{equation}
    \mathcal{L} = \mathcal{L}_{\rm render} + \lambda_{\rm density}\mathcal{L}_{\rm densty} + \lambda_{\rm color}\mathcal{L}_{\rm color},
\end{equation}
where the render loss $\mathcal{L}_{\rm render}$ is given by $\mathcal{L}_{\rm render}=\mathbb{E}_{u,v,s,t}\|C_{u,v,s,t}-C_{gt}\|_2^2$. $\lambda_{\rm density}=10^{-3}$, and $\lambda_{\rm color}=10^{-2}$.

%% file: figures/fig_preliminary.tex
\begin{figure}[bth]
  \includegraphics[width=\linewidth]{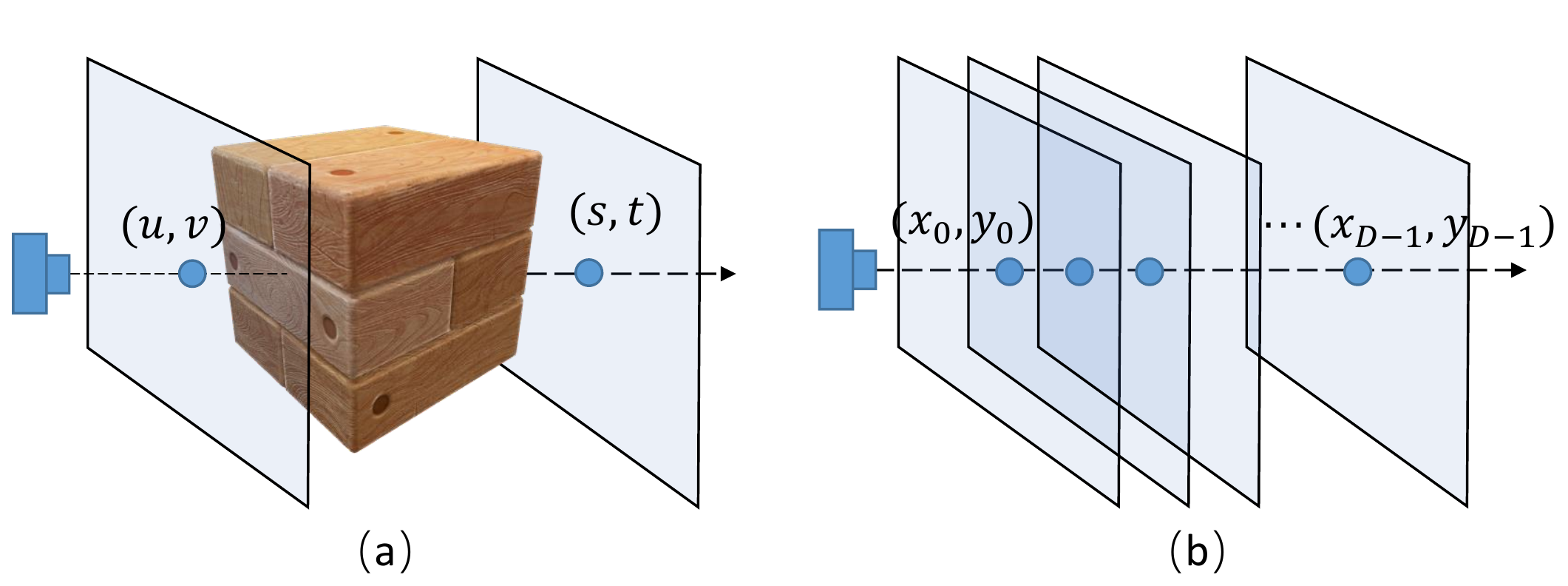}
  \caption{(a) The visualization of a 4D two-plane light field. Each ray is defined by the coordinates of the intersections with the near plane and far plane. (b) We evenly sample planes between the near and the far plane and use the coordinates of intersections on these sample planes as the ray representation.}
  \label{fig:preliminary}
\end{figure}

%% file: figures/fig_arch.tex
\begin{figure*}[bth]
  \includegraphics[width=\linewidth]{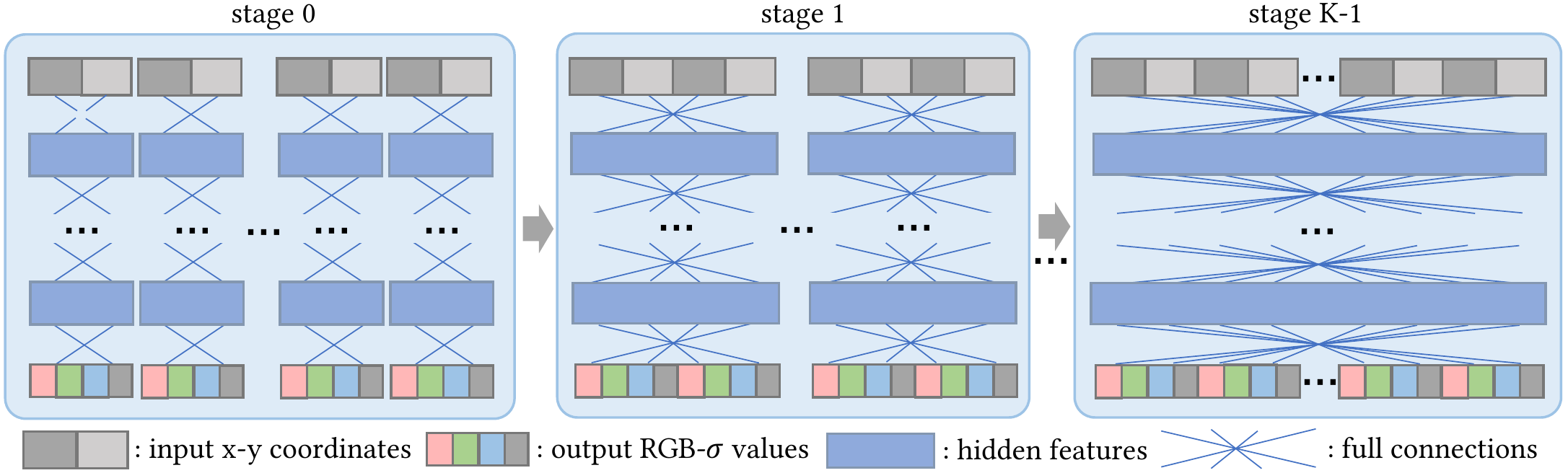}
  \caption{{\bf Progressive training scheme.} In this training scheme, we first separately predict densities and colors of points with different subnetworks, and then we progressively densify the connections between subnetworks to merge them. At the last training stage, we obtain a single fully-connected MLP to predict all the densities and colors of point samples.}
  \label{fig:arch}
\end{figure*}

%% file: figures/fig_compare_llff.tex
\begin{figure*}[!t]
  \includegraphics[width=\linewidth]{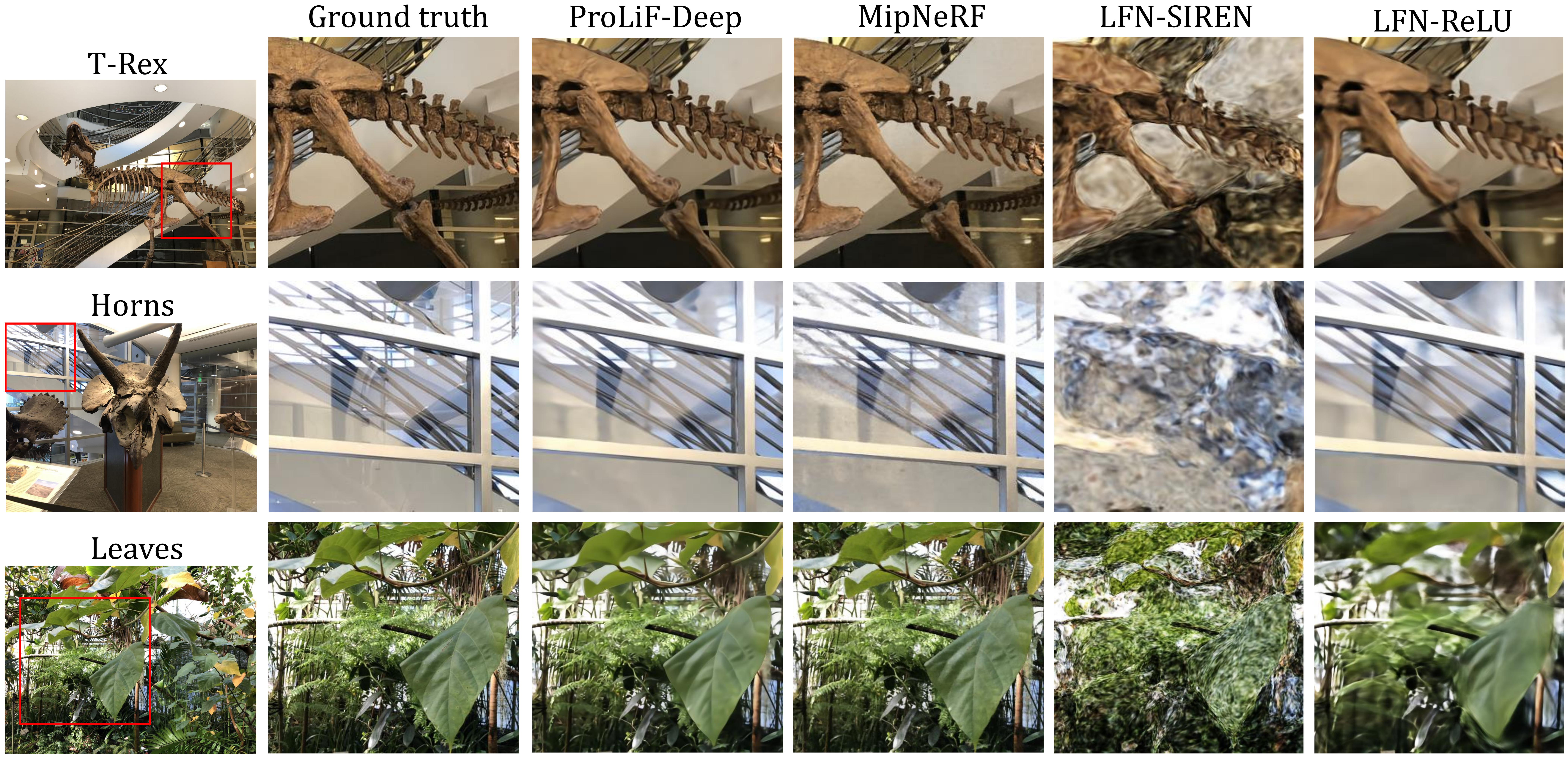}
  \caption{{\bf Qualitative comparison on LLFF dataset.} LFN-SIREN suffers from overtting and LFN-ReLU cannot render fine details. Our method and MipNeRF show better image quality than the two baseline neural light fields.}
  \label{fig:compare_llff}
\end{figure*}

%% file: 4_experiments.tex
\section{Experiments}

\input{tables/quantitative_llff}

\input{tables/quantitative_shiny}

\input{figures/fig_compare_shiny}

\subsection{Experimental Settings}
\noindent {\bf Datasets.} We evaluate ProLiF on two public forward-facing datasets: the LLFF dataset~\cite{mildenhall2019llff} and Shiny dataset~\cite{wizadwongsa2021nex}. Both the datasets contain eight sets of images capturing the real scenes.
The LLFF dataset contains scenes with complex geometries, and the Shiny dataset contains challenging view-dependent effects, like specular highlights and reflections.
The original image resolution of LLFF dataset is $4032\times 3024$, we follow NeRF~\cite{mildenhall2020nerf} and down-sample the images to $1008\times 756$ to fairly compare with different methods.
For the images of Shiny dataset, we follow Nex~\cite{wizadwongsa2021nex} and use the resolution of either $1008 \times 756$ or $1008 \times 567$, depending on the original aspect ratios. For both dataset we choose one out of eight images as test set and the rest as training set. Besides, to demonstrate that our method can be supervised with patch-level loss, like LPIPS~\cite{zhang2018perceptual}, we additionally self-collect a dataset of eight scenes, where each scene is captured under different lighting conditions.

\noindent {\bf Baselines.} We compare our method with the baseline 4D light field networks with different activations of SIREN~\cite{sitzmann2020implicit} or ReLU. 
We also report the comparisons to (1) the state-of-the-art neural volume rendering methods NeRF~\cite{mildenhall2020nerf} and MipNeRF~\cite{barron2021mip}, (2) the state-of-the-art MPI based method Nex~\cite{wizadwongsa2021nex}, and (3) the fast rendering method AutoInt~\cite{lindell2021autoint}. And we additionally report comparisons to two concurrent works (4) ray space embedding network (RSEN)~\cite{attal2021learning} and (5) NeuLF~\cite{liu2021neulf} in the supplementary material for reference.

\noindent {\bf Implementation details.}
The whole training is divided into 5 stages. For k-th training stage (indexed from zero), the number of subnetworks is $16\times 2^{-k}$, and the hidden width of each sub-network is $32\times 2^k$ respectively. Each network is with hidden depth of 8 (for the setting of ProLiF) or 12 (for the setting of ProLiF-Deep), and skip connections are used. We use SIREN activation~\cite{sitzmann2020implicit} with its default parameter $\omega_0=30$ to fit high frequency details. Weight normalization~\cite{salimans2016weight} is used to stabilize the training. The input coordinates are normalized to NDC space~\cite{mildenhall2020nerf}, and at the transition of each training stage, we also subdivide the depth samples. That is, the number of sampled depth planes is $16\times 2^{k}$ at k-th training stage. We train our network on LLFF and Shiny dataset for 1M iterations. In each iteration, we sample 16384 rays for pixel fitting and extra randomly sample 4096 rays for regularization. Adam~\cite{kingma2014adam} optimizer is used with initial learning rate of $1\times 10^{-4}$. The learning rate is gradually decayed to $2.5\times 10^{-6}$ at the end of training using cosine scheduling. The total time of training on a scene is around 28 hours (ProLiF) or 35 hours (ProLiF-Deep) for 1M iterations on a single Nvidia 2080Ti GPU.

\subsection{Comparison}
The quantitative comparisons on the LLFF dataset~\cite{mildenhall2019llff} are presented in Table.~\ref{tab:comparison_llff}. The results show that ProLiF performs better on image quality than the baseline light field networks using SIREN~\cite{sitzmann2020implicit} or ReLU activations. Besides, we achieve comparable rendering quality to the state-of-the-art volume rendering method NeRF~\cite{mildenhall2020nerf} as well as MipNeRF~\cite{barron2021mip}, while ProLiF costs much less time to render an image. We also report the results of the recent fast neural rendering method AutoInt~\cite{lindell2021autoint} and ProLiF shows better rendering quality and computational efficiency.
It is to be noted that to obtain satisfactory results on LLFF dataset, both AutoInt and NeRF need multiple evaluations of the network on sample points on the ray, while our method only needs a single evaluation. Fig~\ref{fig:compare_llff} provides the qualitative results. Baseline light field network with SIREN activation fails to synthesize novel images due to overfitting. Switching to ReLU activation leads to better generalization ability but loses high-frequency details. In comparison, our method and NeRF synthesize novel view images in high quality.

We also report the quantitative comparisons on the Shiny Object dataset~\cite{wizadwongsa2021nex}. This dataset contains eight challenging scenes with complex specular effects and reflections. As shown in Table.~\ref{tab:comparison_shiny}, our method has better rendering quality than NeRF, and is comparable to the MPI based method Nex~\cite{wizadwongsa2021nex}.
Qualitative results in Fig.~\ref{fig:compare_shiny} show that our method has fewer image noises, e.g., noises on the CD and liquid container, than NeRF's render result.

Besides, we report the GPU memory consumption of ProLiF and the SOTA methods when rendering 1024 pixels. As shown in Table.~\ref{tab:gpu_memory}, ProLiF requires much less GPU memory than the other methods, which indicates its excellent compatibility in rendering large patches of pixels.

\input{tables/gpu_memory}

\input{tables/quantitative_ablation}

\subsection{Ablation Studies}
We validate our method designs on the four scenes of the LLFF dataset in Table~\ref{tab:ablation}. Our full model achieves best quantitative results. As shown in Fig~\ref{fig:ablation}, if we directly train the point-based light field network, the results still suffer from overfitting and the underlying scene geometry cannot be correctly learned. After adding the progressive training scheme, the scene geometry can be correctly reconstructed in general but the rendered images are very noisy. Using regularization losses improves the generalization ability but there are observable artifacts if we do not use the progressive training scheme. In contrast, the full model achieves the best quality. 

\input{figures/fig_ablation}

\input{figures/fig_progression}

Besides, we present the qualitative results of our training progression in Fig~\ref{fig:progression}. By increasing the neuron connections in our progressive training scheme, the light field network is able to generate more details on both color images and depth maps.

\input{tables/quantitative_light_change}

\input{figures/fig_light_change}

\subsection{Applications}
\noindent{\bf Rendering under varying light conditions.} To demonstrate the compatibility of our method with image- or patch-level losses, we collect images of eight scenes, where each scene contains images under different light conditions. We change the lighting conditions by controlling the exposure of cameras or switching on/off the external light sources.
To handle different light conditions, we train our network with LPIPS~\cite{zhang2018perceptual} loss, which is a high-level metric that can evaluate the ``perceptual'' similarity between two images, and is robust to noises, intensity changes, distortions, etc. Specifically, in each training step, we randomly render a $300\times 300$ image patch and compare it with the ground-truth image patch to compute the LPIPS loss. Thus, the total training loss is
\begin{equation}
    \mathcal{L} = \mathcal{L}_{\rm reg} + \lambda_{\rm LPIPS}\mathcal{L}_{\rm LPIPS},
\end{equation}
where $\mathcal{L}_{\rm reg}=\lambda_{\rm color}\mathcal{L}_{\rm color} + \lambda_{\rm density}\mathcal{L}_{\rm density}$ is the aforementioned regularization loss, and $\lambda_{\rm LPIPS}$ is set 0.025.
We compare our method with baselines:
(1) NeRF trained with pixel-wise MSE (mean squared error) loss; (2) NeRF trained with LPIPS loss computed from a patch size of $32\times 32$; and 
(3) the NeRF model which renders a $32\times 32$ patch and up-samples the patch to $256\times256$ by a CNN to compute the LPIPS loss. Note that NeRF is not able to efficiently render a large patch in one training step due to the limited computational resources while our method is able to render a large patch of $300\times300$. As shown in Table.~\ref{tab:light}, our method with large-patch LPIPS loss achieves better quality than the two other baselines in terms of SSIM and LPIPS. We do not compare the PSNR since this pixel-wise metric is not compatible with varying light conditions. Qualitative results in Fig~\ref{fig:light} show that NeRF with MSE loss suffers from severe ``foggy'' effects. NeRF with upsampling cannot reconstruct the correct appearance details. NeRF with small-patch LPIPS loss performs better than the original NeRF and that with upsampling but there are still visible artifacts.

\input{figures/fig_clip}
\noindent{\bf Text-guided scene style editing.} We also show that our method can be supervised with CLIP loss~\cite{radford2021learning}. CLIP (Contrastive Language-Image Pre-training) is a large model that is trained on the large scale images-text pairs.
Here we apply CLIP to control the style of the scene with given texts. To achieve this, we introduce appearance embeddings (feature codes), and concatenate the embeddings to the intermediate features of ProLiF to generate images of different appearances. To train it, we propose a loss
\begin{equation}
\mathcal{L}=\mathcal{L}_{\rm render}+\mathcal{L}_{\rm reg}+\lambda_{\rm CLIP}\mathcal{L}_{\rm CLIP},
\end{equation}
where $\mathcal{L}_{\rm render}$ is the MSE loss between the rendered pixels and reference image pixels; $\mathcal{L}_{\rm CLIP}=1-\frac{\mathbf{f}_{\rm image}\cdot \mathbf{f}_{\rm text}}{\|\mathbf{f}_{\rm image}\| \cdot \|\mathbf{f}_{\rm text}\|}$ denotes the semantic distance between the given text and our rendered patches, and $\mathbf{f}_{\rm image}$ and $\mathbf{f}_{\rm text}$ are the extracted features of the rendered image and the given text respectively by the CLIP model. $\lambda_{\rm CLIP}$ is set to 0.05. It is noted that $\mathcal{L}_{\rm CLIP}$ and $\mathcal{L}_{\rm render}$ are calculated from two different appearance embeddings. In each training step we randomly sample 16384 pixels to compute $\mathcal{L}_{render}$, a patch of size $224\times 224$ for $\mathcal{L}_{CLIP}$, and extra 4096 pixels for $\mathcal{L}_{reg}$.

Fig~\ref{fig:clip} shows the qualitative results and we provide videos in the supplementary material. Note our style editing is conducted for the \textbf{whole scene} so that our model is able to produce multi-view consistent images of the same style in the scene. In comparison, previous style editing~\cite{gatys2016image} is only targeted for a \textbf{single image}.

\subsection{Limitations}
Our method is built on the near-far-plane setting and can not be used for 360$^\circ$ scenes. Using the Plucker coordinate~\cite{sitzmann2021light} for a more flexible ray parameterization is an interesting future work. Though our method achieves fast rendering in the inference stage, training such a network for each scene to achieve comparable results to NeRF is still time-consuming. In the future, we may try to decrease the training time by introducing extra data structures~\cite{muller2022instant,yu2021plenoxels,sun2021direct} or explore the possibility of generalizable light fields~\cite{wang2021ibrnet,liu2021neural,yu2021pixelnerf} without the requirement of scene-specific training.

%% file: tables/quantitative_llff.tex
\begin{table*}[!t]
    \begin{tabular}{c||c|c|c|c|c|c|c|c|c||c||c}
        \hline
        Method & Room & Fern & Leaves & Fortress & Orchids & Flower & T-Rex & Horns & Mean & FPS & Model size \\
        \hline
        LFN-SIREN
        & 18.30 & 17.22 & 13.27 & 20.32 & 13.02 & 21.43 & 16.87 & 16.03 & 17.06 & \underline{2.35} & \underline{8.0MB} \\
        LFN-ReLU
        & 27.10 & 20.84 & 17.40 & 28.01 & 15.47 & 24.64 & 22.84 & 23.98 & 22.54 & {\bf 2.90} & {\bf 7.1MB} \\
        ProLiF 
        & \underline{31.49} & \underline{22.91} & \underline{19.94} & \underline{30.23} & \underline{18.62} & \underline{28.08} & \underline{25.84} & \underline{26.86} & \underline{25.50} & 1.40 & 11.0MB \\
        ProLiF-Deep
        & {\bf 31.76} & {\bf 23.03} & {\bf 20.04} & {\bf 30.38} & {\bf 18.82} & {\bf 28.26} & {\bf 26.32} & {\bf 27.34} & {\bf 25.74} & 1.15 & 16.1MB  \\
        \hline
        MipNeRF
        & \underline{32.52} & \underline{25.12} & \underline{20.97} & {\bf 31.42} & \underline{20.28} & {\bf 27.79} & {\bf 27.16} & {\bf 27.55} & {\bf 26.60} & 0.02 & {\bf 2.3MB} \\
        NeRF
        & {\bf 32.70} & {\bf 25.17} & {\bf 20.92} & \underline{31.16} & {\bf 20.36} & \underline{27.40} & \underline{26.80} & \underline{27.45} & \underline{26.50} & \underline{0.04} & \underline{4.4MB}\\
        AutoInt
        & 27.98 & 21.59 & 19.78 & 28.99 & 16.35 & 26.86 & 23.20 & 21.42 & 23.27 & {\bf 0.09} & 5.4MB \\
        \hline
    \end{tabular}
    \caption{{\bf Quantitative results on LLFF dataset.} The top block shows the quantitative results among the neural light field methods,
    and the bottom block additionally reports the quantitative results of NeRF~\cite{mildenhall2020nerf}, MipNeRF~\cite{barron2021mip} and AutoInt~\cite{lindell2021autoint} for reference. ProLiF is much faster than (Mip)NeRF while achieves comparable rendering quality. The numbers with {\bf bold} and \underline{underline} denote the best and the second best respectively. FPS numbers are computed on the resolution of 1008 $\times$ 756. The FPS and model sizes are all computed with the model in the training stage.}
    \label{tab:comparison_llff}
\end{table*}

%% file: tables/quantitative_shiny.tex
\begin{table*}[!t]
    \begin{tabular}{c||c|c|c|c|c|c|c|c|c||c||c}
        \hline
        Method & CD & Tools & Crest & Seasoning & Food & Giants & Lab & Pasta & Mean & FPS & Model size \\
        \hline
        NeRF &
        30.14 & \underline{27.54} & \underline{20.30} & 27.79 & \underline{23.32} & 24.86 & 29.60 & \underline{21.23} & 25.60 & 0.04 & {\bf 4.4MB} \\
        Nex &
        \underline{31.43} & {\bf 28.16} & {\bf 21.23} & {\bf 28.60} & {\bf 23.68} & {\bf 26.00} & 30.43 & {\bf 22.07} & {\bf 26.45} & 0.02 & 401.6MB \\
        ProLiF &
        31.10 & 26.98 & 19.69 & 28.24 & 23.02 & 25.63 & \underline{30.97} & 21.16 & 25.85 & {\bf 1.40} & \underline{11.0MB} \\
        ProLiF-Deep &
        {\bf 31.56} & 27.21 & 19.83 & \underline{28.43} & 23.14 & \underline{25.96} & {\bf 31.26} & \underline{21.23} & \underline{26.07} & \underline{1.15} & 16.1MB \\
        \hline
    \end{tabular}
    \caption{{\bf Quantitative results on Shiny Object dataset.} We report the results in terms of PSNR. ProLiF outperforms NeRF~\cite{mildenhall2020nerf} on speed and image quality, and costs much less memory and time than Nex~\cite{wizadwongsa2021nex}. The rendering time and model sizes are all computed with the model in the training stage.}
    \label{tab:comparison_shiny}
\end{table*}

%% file: figures/fig_compare_shiny.tex
\begin{figure}[bht]
  \includegraphics[width=\linewidth]{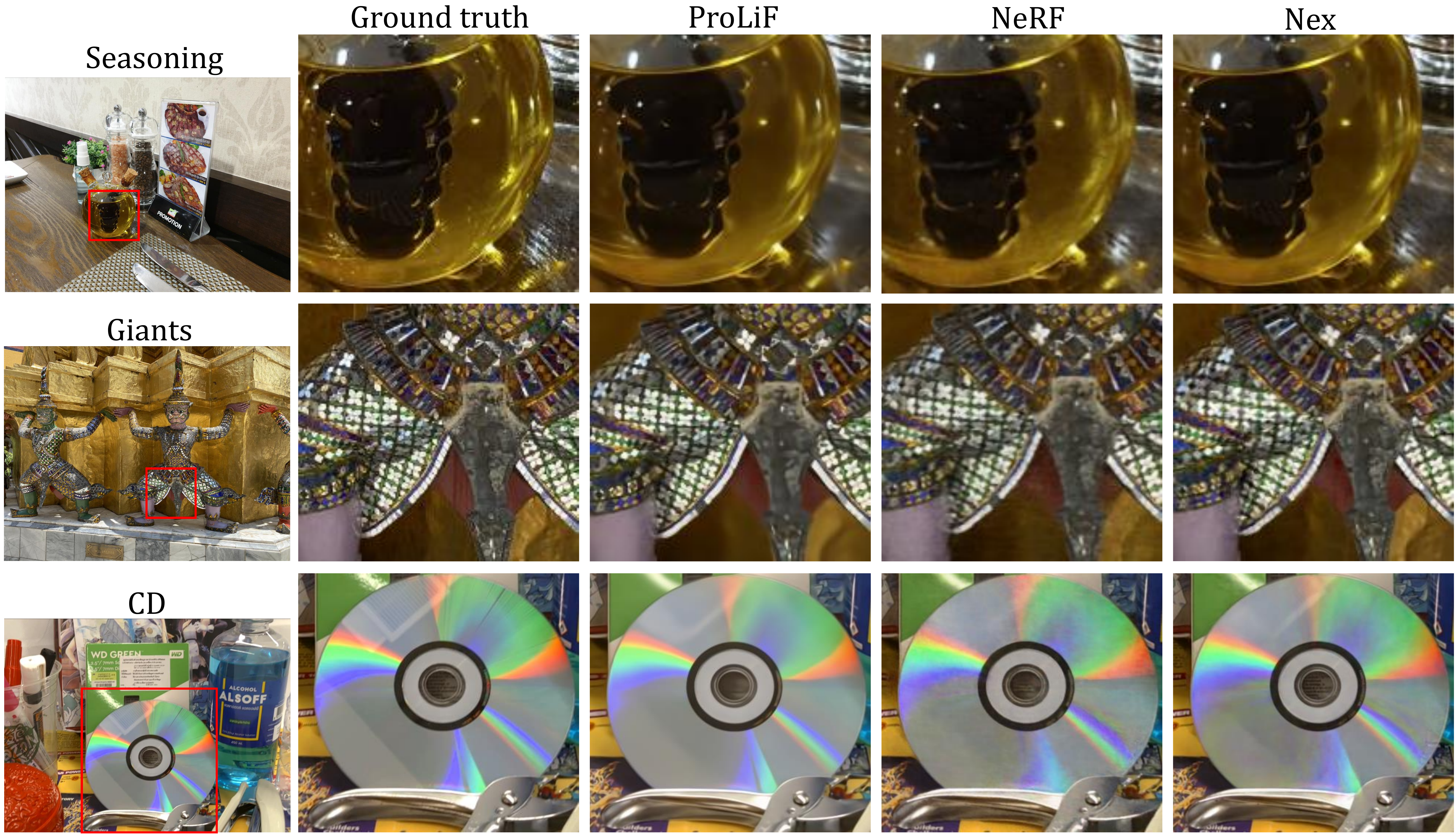}
  \caption{{\bf Qualitative comparison on Shiny Object dataset.} Our method succeeds in recovering speculars, and shows fewer image noises than NeRF.}
  \label{fig:compare_shiny}
\end{figure}

%% file: tables/gpu_memory.tex
\begin{table}[!b]
    \begin{tabular}{c||c|c|c|c||c}
       \hline
       & ProLiF & ProLiF-Deep & NeRF & Nex & AutoInt \\
       \hline
      Mem. & {\bf 105MB} & \underline{148MB} & 1.5GB & 5.1GB & 9.0GB \\
      \hline
    \end{tabular}
    \caption{We provide the GPU memory consumption of the methods to forward 1024 pixels.}
    \label{tab:gpu_memory}
\end{table}

%% file: tables/quantitative_ablation.tex
\begin{table}[!b]
    \begin{tabular}{c||c|c|c|c||c}
       \hline
       & Leaves & Fortress & Trex & Horns & Mean \\
       \hline
      LFN-SIREN
      & 13.27 & 20.32 & 16.87 & 16.03 & 16.62 \\
      Direct training
      & 16.89 & 25.03 & 22.84 & 22.47 & 21.81 \\
      w/o Reg.
      & 16.37 & 25.11 & 22.41 & 22.73 & 21.65 \\
      w/o Prog.
      & 19.33 & 29.56 & 25.35 & 25.37 & 24.91 \\
      Full method
      & {\bf 19.94} & {\bf 30.23} & {\bf 25.84} & {\bf 26.86} & {\bf 25.72} \\
      \hline
    \end{tabular}
    \caption{{\bf Ablation study.} We provide quantitative results in terms of PSNR on four scenes of the LLFF dataset.}
    \label{tab:ablation}
\end{table}

%% file: figures/fig_ablation.tex
\begin{figure}[!t]
  \includegraphics[width=\linewidth]{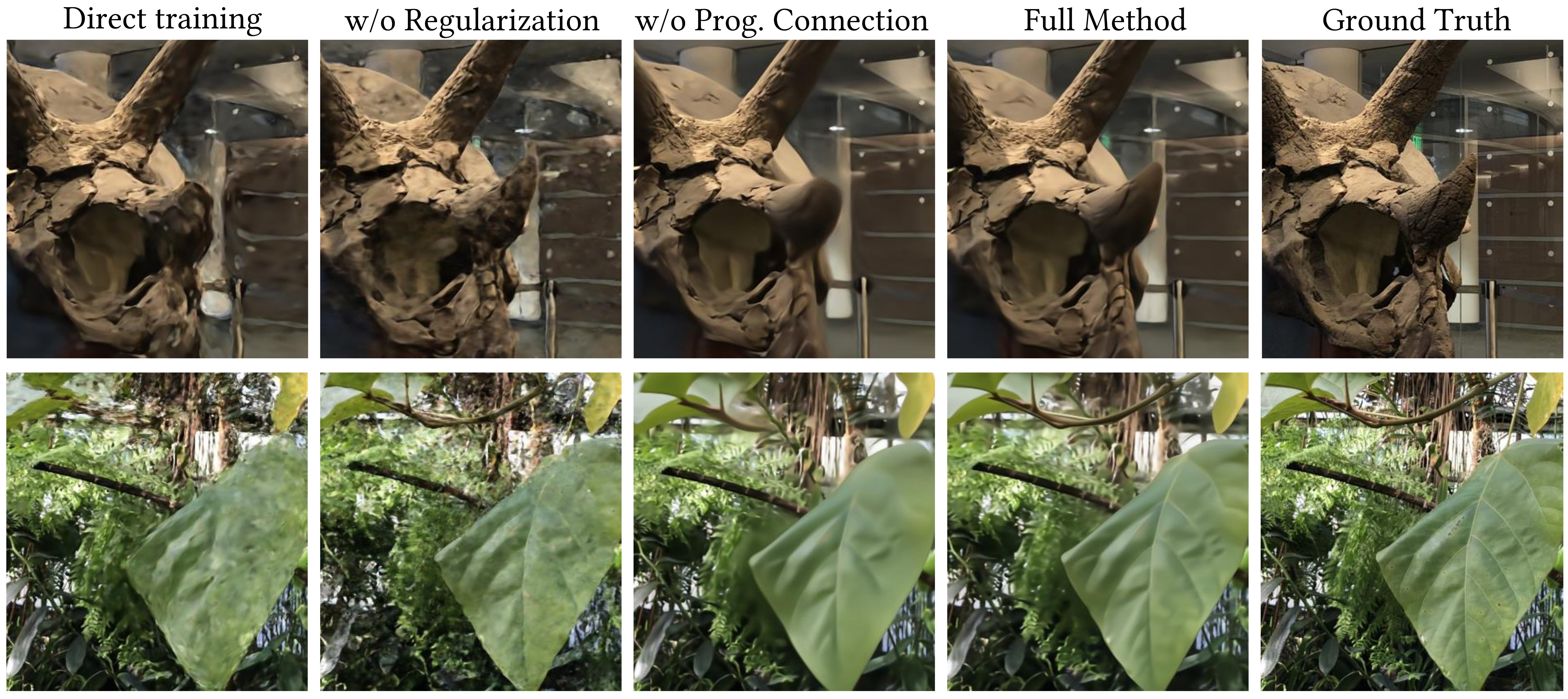}
  \caption{{\bf Ablation study.} We present the visual comparison with different method designs.}
  \label{fig:ablation}
\end{figure}

%% file: figures/fig_progression.tex
\begin{figure}[!t]
  \includegraphics[width=\linewidth]{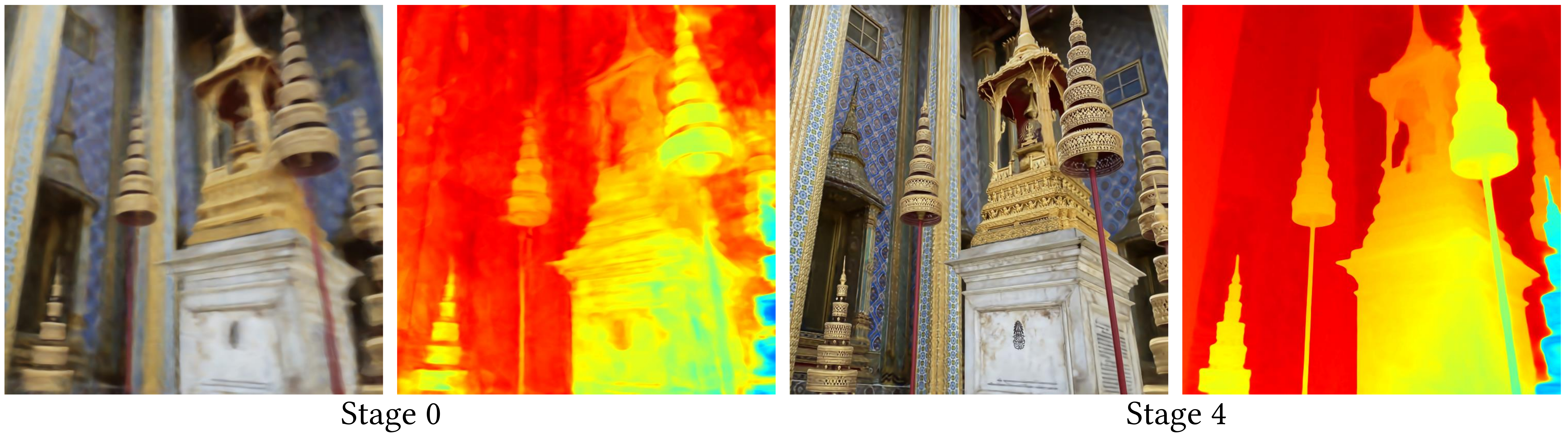}
  \caption{{\bf Training progression.} We show the rendered novel views and depth maps of ProLiF at the end of the first stage and the last stage.}
  \label{fig:progression}
\end{figure}

%% file: tables/quantitative_light_change.tex
\begin{table}[!b]
    \begin{tabular}{c||c|c|c|c}
    \hline
     & Fu & Bench & Sign & Coner \\
     \hline
     NeRF
     & .535/.283 & .456/.251 & \textbf{.640}/.190 & .582/.212 \\
     NeRF-P
     & .735/\textbf{.077} & .433/.206 & .527/.179 & .695/.096 \\
     NeRF-U
     & .654/.139 & .266/.257 & .477/.226 & .626/.171 \\
     ProLiF
     & \textbf{.763}/\textbf{.077} &\textbf{ .517}/\textbf{.095} & .613/\textbf{.123} & \textbf{.830}/\textbf{.061} \\
     \hline
     & Station & Totoro & Doll & Stair \\
     \hline
    NeRF
    & .519/.227 & .548/.243 & .729/.173 & .816/.101 \\
    NeRF-P
    & .532/.161 & .694/.124 & \textbf{.785}/\textbf{.098} & \textbf{.866}/\textbf{.045} \\
    NeRF-U
    & .391/.244 & .549/.206 & .640/.190 & .713/.127 \\
    ProLiF
    & \textbf{.679}/\textbf{.109} & \textbf{.741}/\textbf{.115} & .755/\textbf{.098} & .862/.049 \\
    \hline
    \end{tabular}
    \caption{{\bf Quantitative results on the light change dataset.} We report the metrics of SSIM$\uparrow$/LPIPS$\downarrow$ for each scene. `NeRF-P' denotes NeRF with 32 $\times$ 32 patch supervision, and `NeRF-U' denotes NeRF with 256 $\times$ 256 up-sampled patch supervision.}
    \label{tab:light}
\end{table}

%% file: figures/fig_light_change.tex
\begin{figure*}[!t]
  \includegraphics[width=\linewidth]{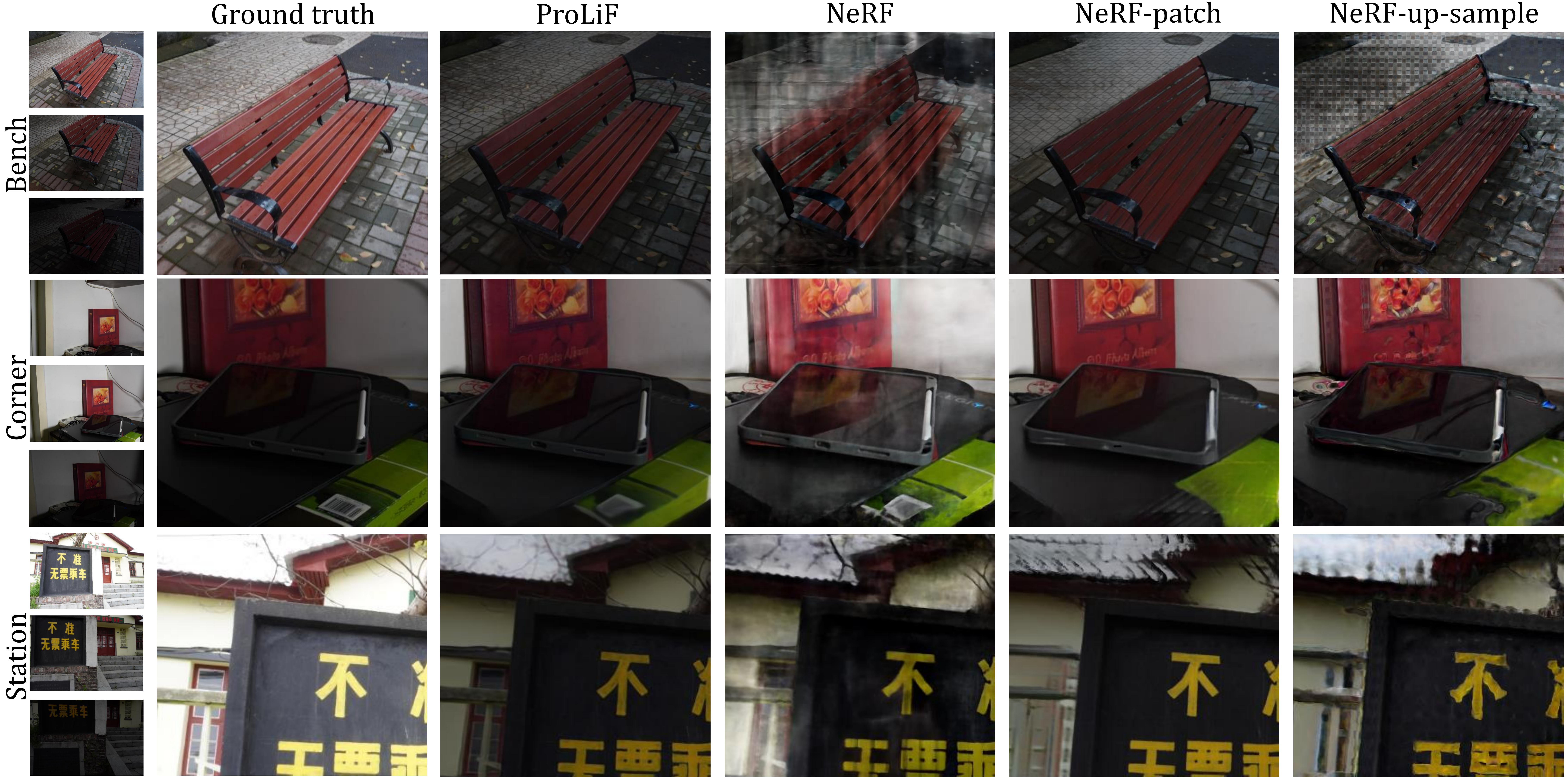}
  \caption{{\bf Novel view synthesis using input images with varying light conditions.} The first column shows a subset of input images while the other columns show the rendered images of different methods. Our method with LPIPS succeeds in recovering the details while achieving robustness to light changes.}
  \label{fig:light}
\end{figure*}

%% file: figures/fig_clip.tex
\begin{figure}[htb]
  \includegraphics[width=\linewidth]{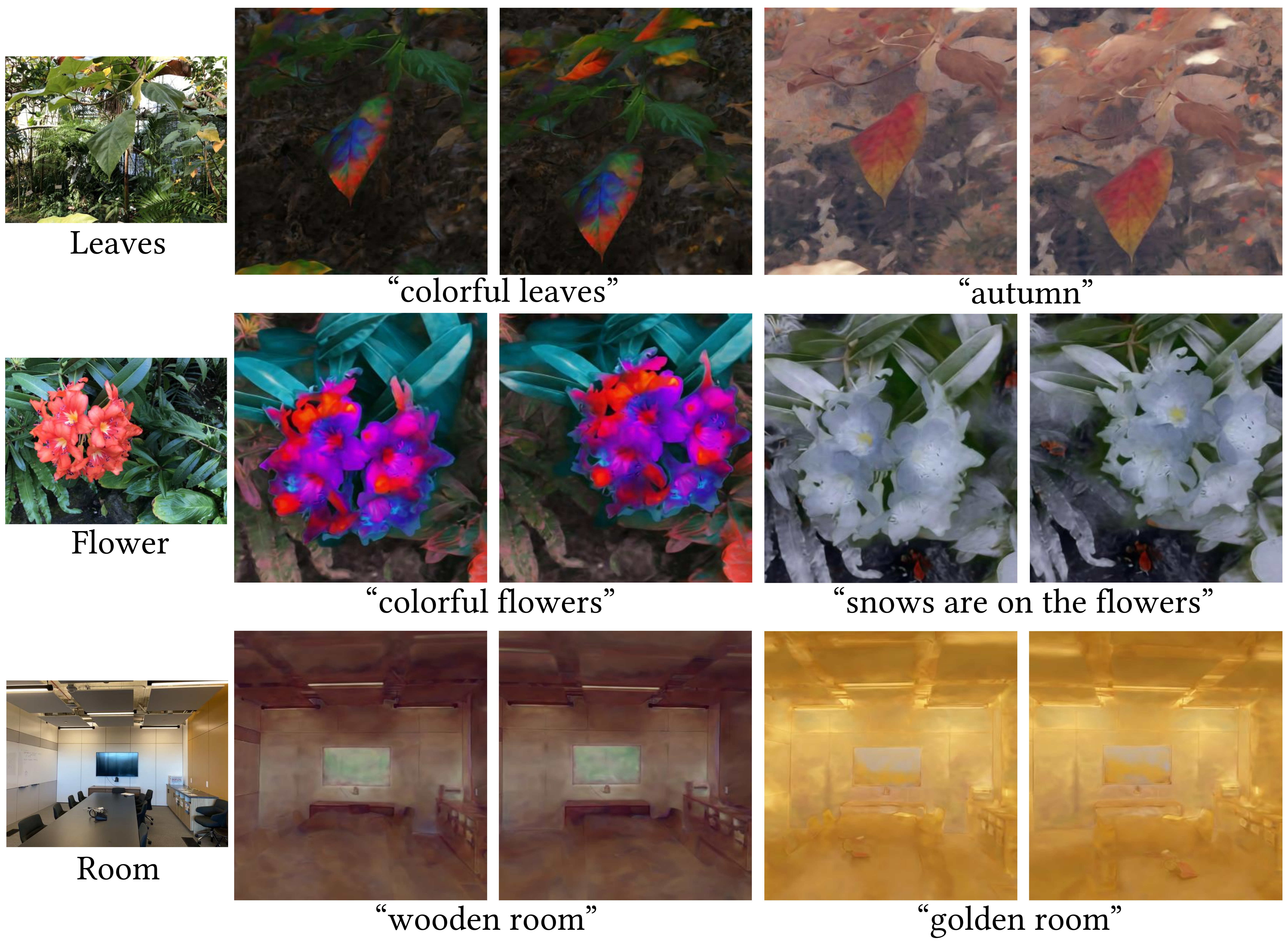}
  \caption{{\bf Examples of text-guided style control using CLIP loss.} First column shows the original image while the other columns show the used texts and the rendered images. Note we manipulate the style of the whole scene so that we are able to render multi-view consistent images of the same style.}
  \label{fig:clip}
\end{figure}

%% file: 5_conclusion.tex
\section{Conclusion}
We have presented ProLiF, a compact neural representation for efficient view synthesis with around 40$\times$ speeding up than NeRF. ProLiF, as a light field network, is able to efficiently render a large patch of pixels in a single training step and therefore has better compatibility with image- or patch-level losses such as LPIPS and CLIP loss. Directly training a light field network often suffers from overfitting, which cannot generalize to novel views. To address this, we propose a novel progressive training scheme and regularization losses to preserve multi-view consistency of rendered images. In the future, we may speed up the training of ProLiF and incorporate it in more applications, such as 3D consistent image generation.

%% file: 6_ack.tex
\begin{acks}
We thank Jiepeng and Xiaoxiao for the proofreading. Lingjie Liu was supported by Lise Meitner Postdoctoral Fellowship. Computational resources are mainly provided by HKU GPU Farm.
\end{acks}

%% file: supp_1_addition_method_descriptions.tex
\section{Additional Method Description}

\subsection{Details of progressive connection}
\vspace{2pt}
\noindent{\bf Architecture.}
We define the architecture of ProLiF at training stage $k$ by $N_k = D\times 2^{-k}$ subnetworks $\Phi_{k}$ that $\Phi_{k} = \{F_k^0, F_k^1,...,F_k^{N_k-1}\}$. Each subnetwork is with hidden width $W\times2^{k}$. Here $D$ and $W$ are preset parameters. Note that the sum of hidden widths for each layer is constant during the whole training progress.
For each subnetwork, its input is a concatenation of the sampled x-y coordinates, and the output is the corresponding concatenated color and density values, as shown in Fig.~4 of the main paper. When $k=\log_2(D)$ at the last training stage, ProLiF is a single fully-connected MLP that $\Phi_k = \{ F_k^0 \}$.


\vspace{2pt}
\noindent{\bf Add connections.}
We now consider to increase the network capacity by merging the subnetworks at the end of k-th training stage. Specifically, we merge the neighboring sub-networks $F_k^{2i}$ and $F_k^{2i+1}$ to $F_{k+1}^{i}$ for $i \in \{0, ..., D\times 2^{-(k+1)} - 1\}$. After merging, we expect the output values of the whole network $\Phi_{k}$ are kept unchanged. To achieve this, we consider constructing the weight matrix $W_{k+1}^{i}$ and bias vector $b_{k+1}^i$ of each layer function $z_{k+1}^{i} = \omega(W_{k+1}^{i}y_{k+1}^{i} + b_{k+1}^i)$ of the MLP, where $y$ and $z$ is the input and output of the layer respectively, and $\omega$ is either an activation function (for the first and hidden layers) or the identity function (for the last layer). Given the concatenation of the layer input $y_{k+1}^{i} = \left[y_k^{2i}, y_k^{2i+1}\right]^\top$, we expect the output after connection equals the corresponding concatenated outputs, i.e., we expect $z_{k+1}^{i} = \left[z_k^{2i}, z_k^{2i+1}\right]^\top$. To this end, we construct the weight matrix and bias vector as
\begin{equation}
    W_{k+1}^{i} = \begin{bmatrix} W_{k}^{2i} & 0 \\ 0 & w_{k}^{2i+1} \end{bmatrix}, b_{k+1}^{i} = \begin{bmatrix} b_{k}^{2i} \\ b_{k}^{2i+1} \end{bmatrix}.
\end{equation}
This is also described by the Python-style code snippet that
\begin{lstlisting}[language=Python]
def add_connections(org_weight, org_bias):
    # org_weight - shape of [n_subnetworks, d_out, d_in]
    # org_bias   - shape of [n_subnetworks, d_out]
    n_subnetworks, d_out, d_in = org_weight.shape
    weight = zeros(n_subnetworks//2, d_out*2, d_in*2)
    weight[:, :d_out, :d_in] = org_weight[0::2, :, :]
    weight[:, d_out:, d_in:] = org_weight[1::2, :, :]
    bias = rearrange(org_bias, '(n c) o -> n (c o)', c=2)
    return weight, bias
\end{lstlisting}


\vspace{2pt}
\noindent{\bf Subdivide depth samples.}
When switching between the training stages, we also subdivide the depth planes for better rendering quality. Therefore, the number of depth samples at training stage $k$ is $D_k = D\times 2^{k}$.
Similar to progressive connection, the rendered color of each input ray is kept unchanged after we perform the subdivision. We achieve this by reparameterizing the weight matrices and bias vectors of the first and the last layer functions of the networks. 

\vspace{2pt}
\noindent{\it (a) Subdivide input depth samples.}  Given the ray parameters $(u, v, s, t)$, the sampled x-y coordinates $\left\{x_k^0, y_k^0, ..., x_k^{D_k - 1}, y_k^{D_k - 1} \right\}$ are computed by
\begin{equation}
\begin{split}
    x_k^i = u + (s - u)d_k^i, \\
    y_k^i = v + (t - v)d_k^i.
\end{split}
\end{equation}
We make the concatenation of x-y coordinates as the network input. At the end of k-th training stage, the network input is changed from 
\[
p_k = \left[x_k^0, y_k^1, ..., x_k^{D_k - 1}, y_k^{D_k - 1} \right]^{\top}
\]
to
\[
p_{k+1} = \left[x_{k+1}^0, y_{k+1}^1,..., x_{k+1}^{D_{k+1} - 1}, y_{k+1}^{D_{k+1} - 1} \right]^\top.
\]
The sampled depth values at stage $k$ are defined by $\{d_k^i = \frac{i}{D_k} + \frac{1}{2D_k} | i=0,...,D_k-1\}$. It is to be noted that we assume the near plane and far plane are placed at depth value of $0$ and $1$. Then it is not difficult to find that $x_{k}^{i} = (x_{k+1}^{2i} + x_{k+1}^{2i+1})/2$ and $y_{k}^{i} = (y_{k+1}^{2i} + y_{k+1}^{2i+1})/2$.
With this observation we can construct the new weight and bias parameters of the first layer, which is described by the code snippet that
\begin{lstlisting}[language=Python]
def subdiv_input_layers(org_weight, org_bias):
    # org_weight - shape of [n_subnetworks, d_out, d_in]
    # org_bias   - shape of [n_subnetworks, d_out]
    bias = org_bias
    weight = rearrange(org_weight, 'n o (d c) -> n o d c', c=2)
    weight = concatenate([weight, weight], dim=3) * 0.5
    weight = rearrange(weight, 'n o d c -> n o (d c)')
    return weight, bias
\end{lstlisting}
It can be validated that this construction of new weight and bias parameters will not affect the values of hidden features after depth subdivision.

\noindent{\it (b) Subdivide output predictions.} We also subdivide the output color and density samples by reparameterizing the last layer of each subnetwork.
To obey the principle that the final rendered colors will not be affected after the reparameterization, we double-copy the original parameters of the last layers, which is described by the code that
\begin{lstlisting}[language=Python]
def subdiv_output_layers(org_weight, org_bias):
    # org_weight - shape of [n_subnetworks, d_out, d_in]
    # org_bias   - shape of [n_subnetworks, d_out]
    bias = rearrange(org_bias, 'n (d c) -> n d c', c=4)
    bias = concatenate([bias, bias], dim=2)
    bias = rearrange(bias, 'n d c -> n (d c)')
    weight = rearrange(org_weight, 'n (d c) i -> n d c i', c=2)
    weight = concatenate([weight, weight], dim=2)
    weight = rearrange(weight, 'n d c i -> n (d c) i')
    return weight, bias
\end{lstlisting}

\subsection{Derivation of consistency losses}
In this section we derive the formula of consistency losses for regularization.

\vspace{2pt}
\noindent{\bf Density consistency loss.}
As described in the main paper, we denote the whole process of computing a density $\sigma_i$ on a point $p_i=(x_i, y_i)$ at depth $d_i$ on the ray $(u, v, s, t)$ as $\sigma_i=L_i(u, v, s, t)$.
The derivation consists of two parts: 1) we derive that all rays intersecting the same point form an affine space, and 2) we perform locality analysis on the derived affine space.

1) For an arbitrary ray $(u', v', s', t')$ that intersects $p_i$, we have
\begin{equation}
\begin{cases}
x_i &= u' + (s' - u') d_i \\
y_i &= v' + (t' - v') d_i. \\
\end{cases}
\end{equation}
We already know that $(u,v,s,t)$ intersects on $p_i$ so we have
\begin{equation}
\begin{cases}
x_i &= u + (s - u) d_i \\
y_i &= v + (t - v) d_i. \\
\end{cases}
\end{equation}
Then, we have
\begin{equation}
\begin{split}
\begin{cases}
    u' + (s' - u') d_i &= u + (s - u) d_i \\
    v' + (t' - v') d_i &= v + (t - v) d_i
\end{cases}
\\
\Rightarrow
\begin{cases}
-(u' - u) * (1 - d_i) &= (s' - s) * d_i \\
-(v' - v) * (1 - d_i) &= (t' - t) * d_i.
\end{cases}
\end{split}
\label{eqn:manifold}
\end{equation}
Eqn~\ref{eqn:manifold} denotes that all intersected ray $(u', v', s', t')$ can be represented by two parameters $\alpha$, $\beta$ that
\begin{equation}
(u',v',s',t')^\top = (u,v,s,t)^\top + \alpha\mathbf{p} + \beta\mathbf{q},
\end{equation}
where
\begin{equation}
\begin{split}
    \mathbf{p} &= (d_i, 0, -(1 - d_i), 0)^{\top}, \\
    \mathbf{q} &= (0, d_i, 0, -(1 - d_i))^{\top}.
\end{split}
\label{eqn:def_p_q}
\end{equation}
This is actually an affine space with a fixed point $(u,v,s,t)$ and spanning basis $\mathbf{p}$ and $\mathbf{q}$. 

2) We now perform local analysis of the density value around $(u, v, s, t)$. For a first-order approximation of the density value at the neighboring ray $(u',v',s',t')$, we have
\begin{equation}
\begin{split}
    L_i(u', v', s', t') &\approx L_i(u, v, s, t) + \nabla L_i\cdot(\alpha\mathbf{p} + \beta\mathbf{q}) \\
    &\approx L_i(u, v, s, t) + \alpha\nabla L_i\cdot\mathbf{p} + \beta\nabla L_i\cdot\mathbf{q}.
\end{split}
\end{equation}
To regularize the density value, we expect
\begin{equation}
\begin{split}
L_i(u', v', s', t') - L_i(u, v, s, t) &= 0 \\
\Rightarrow \alpha\nabla L_i\cdot\mathbf{p} + \beta\nabla L_i\cdot\mathbf{q} &\approx 0, 
\end{split}
\end{equation}
which is a linear combination of two projected values.
Note that $\alpha$ and $\beta$ can be arbitrary combinations, therefore we directly minimize the individual projected terms. This derives the density consistency loss in Eqn~2 of the main paper.

\vspace{2pt}
\noindent{\bf Color consistency loss.}
The derivation of color consistency loss is the same as the density consistency loss by replacing the density function $L_i$ with the color function $LF$.

Alternatively, we may constraint the sample color $c_i$ instead of the output ray color $C$. In this case, we expect the sampled colors $c_i$ to be consistent on different ray parameters and apply the same loss to the sampled colors as to the densities. However, in our early experiments, we found that imposing such a loss on the sample colors cannot help achieve view-consistent results. We think it is due to the accumulated error - though we regularize every sampled colors, the final rendered pixel color, which is synthesized by volume accumulation from all the sampled colors, still suffers from multi-view inconsistency. Therefore, we instead directly regularize the rendered colors. That is, the final rendered colors of rays intersecting the same surface point are expected to be consistent.

\input{supp_figures/fig_clip_arch}
\subsection{Text-guided scene style editing}
\vspace{2pt}\noindent
{\bf Network architecture.} To edit the style of the scenes by texts using CLIP loss, we choose to use a slightly different architecture to that used for the original NVS task, and in this architecture we introduce the appearance embeddings. The architecture for this task is shown in Fig.~\ref{fig:arch_clip}. For simplicity of description, we only visualize the architecture at the last training stage. As already described in the main paper, the overall training loss is $\mathcal{L}=\mathcal{L}_{\rm render}+\mathcal{L}_{\rm reg}+\lambda_{\rm CLIP}\mathcal{L}_{\rm CLIP}$. And $\mathcal{L}_{\rm CLIP}$ and $\mathcal{L}_{\rm render}$ are calculated from two different appearance embeddings.

\input{supp_tables/compare_nelf}
\input{supp_tables/compare_rsen}

\input{supp_figures/fig_stop_gradient}
\vspace{2pt}\noindent
{\bf Effect of stop-gradient operators.} In the implementation, we add the stop-gradient operators to the output densities and a hidden feature of the appearance network (green dotted arrows in Figure~\ref{fig:arch_clip}) when calculating $\mathcal{L}_{\rm CLIP}$. We use this strategy to regularize the rendered results of style-editing to a similar intrinsic structure of the original scene. The qualitative comparisons between w/ stop-gradient and w/o stop-gradient are presented in Figure~\ref{fig:stop_gradient}.

%% file: supp_figures/fig_clip_arch.tex
\begin{figure}[htb]
  \includegraphics[width=\linewidth]{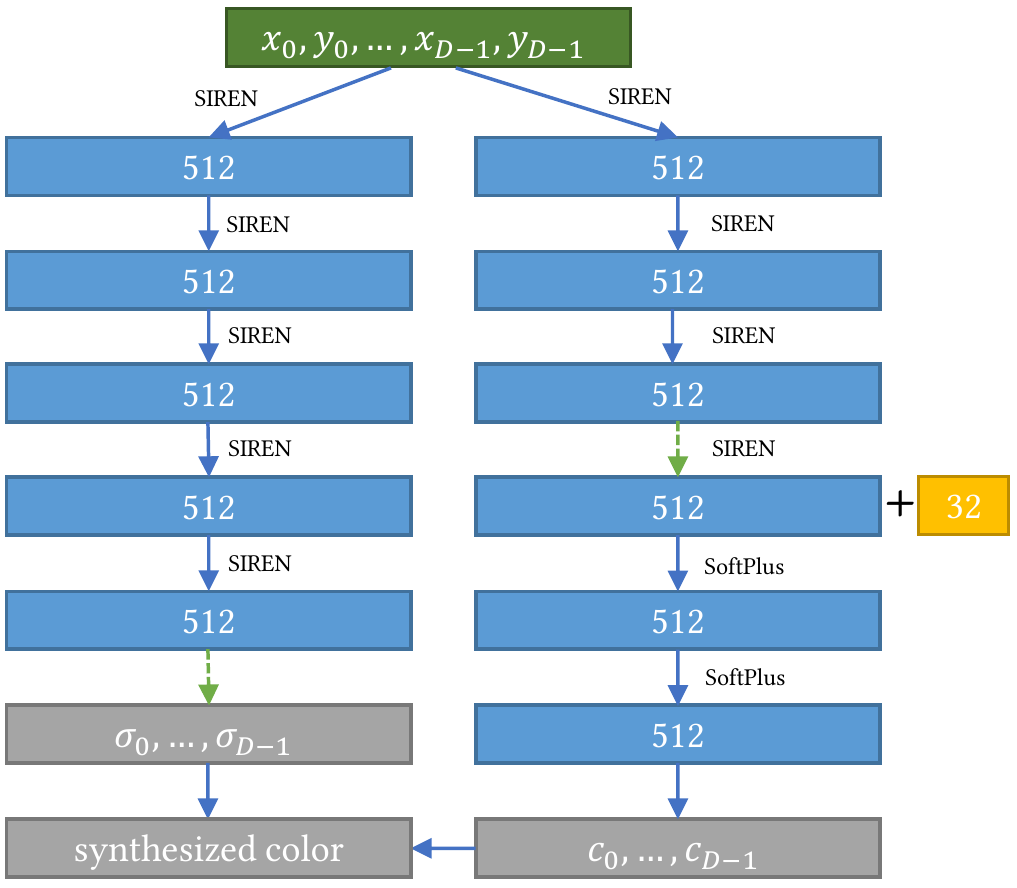}
  \caption{{\bf Architecture for text-guided scene editing.} The yellow block denotes the appearance embedding, and the gray blocks denote the output RGB and density samples, as well as the final rendered colors. The green dotted arrows denote stop-gradient operators when calculating $\mathcal{L}_{\rm CLIP}$.}
  \label{fig:arch_clip}
\end{figure}

%% file: supp_tables/compare_nelf.tex
\begin{table*}[hbt]
    \begin{tabular}{c||c|c|c|c|c|c|c|c|c||c||c}
        \hline
        Method & CD & Tools & Crest & Seasoning & Food & Giants & Lab & Pasta & Mean & FPS & Model size \\
        \hline
        NeuLF &
        {\bf 32.11} & 26.73 & {\bf 20.11} & 27.12 & 22.61 & 24.95 & {\bf 31.95} & 20.64 & 25.78 & {\bf 2.65} & {\bf 5.0MB} \\
        ProLiF &
        31.10 & \underline{26.98} & 19.69 & \underline{28.24} & \underline{23.02} & \underline{25.63} & 30.97 & \underline{21.16} & \underline{25.85} & \underline{1.40} & \underline{11.0MB} \\
        ProLiF-Deep &
        \underline{31.56} & {\bf 27.21} & \underline{19.83} & {\bf 28.43} & {\bf 23.14} & {\bf 25.96} & \underline{31.26} & {\bf 21.23} & {\bf 26.07} & 1.15 & 16.1MB \\
        \hline
    \end{tabular}
    \caption{{\bf Comparison to NeuLF on Shiny Object dataset.} We report the results in terms of PSNR. FPS numbers are computed on the resolution of 1008 $\times$ 756.}
    \label{tab:supp_comparison_shiny}
\end{table*}

%% file: supp_tables/compare_rsen.tex
\begin{table*}[hbt]
    \begin{tabular}{c||c|c|c|c|c|c|c|c|c||c||c}
        \hline
        Method & Room & Fern & Leaves & Fortress & Orchids & Flower & T-Rex & Horns & Mean & FPS & Model size \\
        \hline
        RSEN
        & {\bf 33.57} & 24.25 & {\bf 21.82} & 31.46 & {\bf 20.29} & 28.71 & {\bf 29.41} & {\bf 30.12} & {\bf 27.45} & 0.43 & {\bf 4.8MB} \\
        ProLiF
        & 32.77 & \underline{24.66} & 21.50 & \underline{31.56} & 19.40 & \underline{29.07} & 27.42 & 28.91 & 26.91 & {\bf 5.22} & \underline{11.0MB} \\
        ProLiF-Deep
        & \underline{32.98} & {\bf 24.81} & \underline{21.65} & {\bf 31.63} & \underline{19.62} & {\bf 29.26} & \underline{27.92} & \underline{29.32} & \underline{27.15} & \underline{4.58} & 16.1MB \\
        \hline
        RSEN (with teacher)
        & 34.04 & 26.06 & 22.27 & 32.60 & 21.10 & 28.90 & 28.80 & 29.76 & 27.94 & 0.43 & 4.8MB
    \end{tabular}
    \caption{{\bf Comparison to RSEN on LLFF dataset.} We report the results in terms of PSNR. FPS numbers are computed on the resolution of 504 $\times$ 378. We additionally provide the results of RSEN that uses a NeRF model as teacher for extra supervison, for reference.}
    \label{tab:supp_comparison_llff}
\end{table*}

%% file: supp_figures/fig_stop_gradient.tex
\begin{figure}[htb]
  \includegraphics[width=\linewidth]{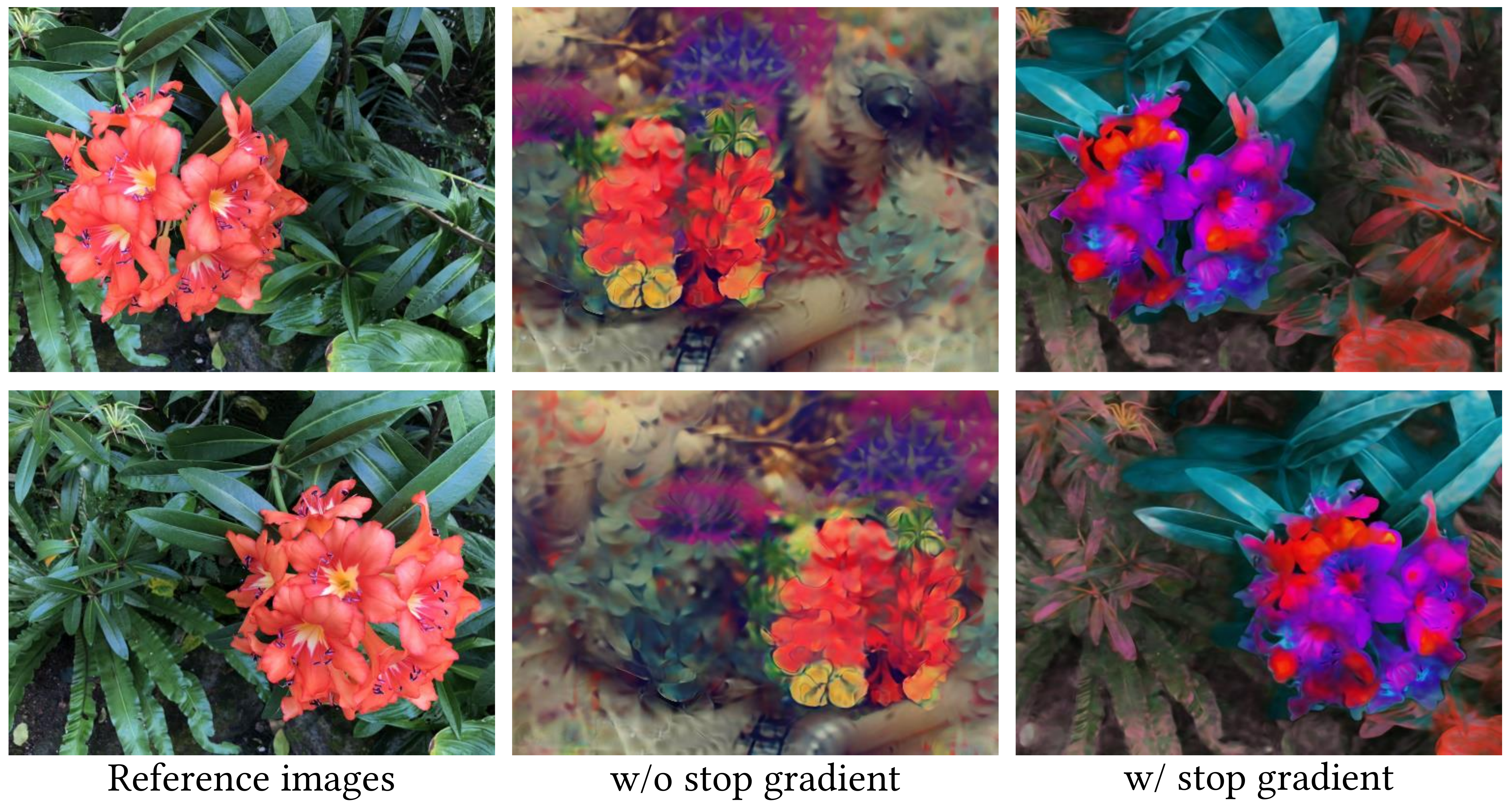}
  \caption{{\bf Effect of stop-gradient operators.} The top and bottom rows show images from two different views. The text for $\mathcal{L}_{\rm CLIP}$ here is "colorful flowers". Adding stop-gradient operators helps the rendered images preserve the original instrinsic structures.}
  \label{fig:stop_gradient}
\end{figure}

%% file: supp_2_additional_results.tex
\section{Additional results}
Here we additionally report the quantitative comparisons to the concurrent works NeuLF~\cite{liu2021neulf} and Ray Space Embedding Networks (RSEN)~\cite{attal2021learning}. Since the codes of these works have not been made public yet, we compare the quantitative results with those reported in their papers. Table~\ref{tab:supp_comparison_shiny} provides the comparisons to NeuLF on the Shiny Object dataset~\cite{wizadwongsa2021nex}, where our method performs better rendering quality than NeuLF. The results of NeuLF on the LLFF dataset are not reported in their paper.

Table~\ref{tab:supp_comparison_llff} shows the comparisons to RSEN. The experiments of RSEN on LLFF dataset are conducted on the resolution of $504\times378$, and we compare to it using the same resolution. Our method achieves comparable results as RSEN while is much more efficient in rendering.
